\documentclass[journal]{IEEEtran}
\usepackage{amsfonts}
\usepackage{subfigure}
\usepackage{mathrsfs}
\usepackage{float}
\usepackage{graphics}
\usepackage{epstopdf}
\usepackage{times}
\usepackage{booktabs}
\usepackage{amsmath}
\usepackage{amssymb}
\usepackage{indentfirst,bm}
\usepackage[dvips]{graphicx}
\usepackage{epsfig}
\usepackage{epsf}
\usepackage{color}
\usepackage{cite}
\usepackage{enumerate}
\usepackage{multirow}
\usepackage{stfloats}
\usepackage{amsmath}
\usepackage{tikz}
\usetikzlibrary{positioning, fit, calc, shapes.geometric, arrows.meta}
\usepackage{caption}
\usepackage{algorithm}
\usepackage{algorithmicx}
\usepackage{algpseudocode,lineno}

\usepackage{hyperref}

\usepackage[compact]{titlesec} 
\titlespacing*{\section}{0pt}{*1.3}{*0.8}
\titlespacing*{\subsection}{0pt}{*1.2}{*0.7}

\usepackage{tikz}
\usetikzlibrary{shapes.geometric, arrows}

\tikzstyle{startstop} = [rectangle, rounded corners, minimum width=2.5cm, minimum height=1cm, text centered, draw=black, fill=blue!30]
\tikzstyle{process} = [rectangle, minimum width=2.5cm, minimum height=1cm, text centered, draw=black, fill=orange!30]
\tikzstyle{decision} = [diamond, aspect=2, text centered, draw=black, fill=yellow!30]
\tikzstyle{arrow} = [thick,->,>=stealth]
\usepackage{amsthm}

\newtheoremstyle{mydefstyle}
  {0.5em} 
  {0.5em} 
  {\itshape} 
  {} 
  {\bfseries} 
  {.} 
  { } 
  {} 

\theoremstyle{mydefstyle}

\begin{document}

\bstctlcite{BSTcontrol}

\title{High-Order Liquid Evidence Encoding for Gradual GNSS Spoofing Detection in Autonomous Driving}

\author{Muhammad Ayub Sabir, Junbiao~Pang, Fatima~Ashraf
\IEEEcompsocitemizethanks{

\IEEEcompsocthanksitem Sabir, J. Pang, and Fatima are with the Faculty of Information Technology, Beijing University of Technology, Beijing 100124, China (e-mail: sabir@emails.bjut.edu.cn; junbiao\_pang@bjut.edu.cn;
fatimaashraf@emails.bjut.edu.cn)

 }
}

\maketitle


\begin{abstract}
Accurate Global Navigation Satellite System (GNSS)-based localization is essential for safe and reliable autonomous driving. However, spoofing attacks can manipulate vehicle position estimates. Continuous and subtle attacks are particularly difficult to detect because individual GNSS observations may remain plausible while the inconsistency between GNSS-implied displacement and onboard vehicle motion gradually increases. Existing methods often rely on static vehicle-behavior features or a single residual signal and do not explicitly model this evolution. To address this problem, we propose a causal high-order liquid evidence framework for GNSS spoofing detection. The method first constructs a physics-guided GNSS--motion inconsistency residual by comparing GNSS-implied displacement with onboard-motion-derived displacement. It then forms separate evidence streams for the residual level and its first- and second-order discrete variations, with relevant contextual cues selected according to the evidence order. Each stream is processed by a separate adaptive liquid encoder, and the resulting temporal states are hierarchically coupled to predict spoofing at the window endpoint
using only current and past observations. Experiments on three subsets of the real-world AV-GPS dataset show that the proposed method achieves the highest F1-scores among the evaluated temporal models on Dataset~1 and Dataset~3, reaching 0.9535 and 0.9777, respectively. On Dataset~3, it detects both labeled normal-to-attack transitions within four sampling steps. Code and datasets are publicly available at:\url{https://github.com/pangjunbiao/GNSS_Spoofing.git}.
\end{abstract}


\section{Introduction}
\label{sec:introduction}

Autonomous vehicles increasingly rely on Global Navigation Satellite System (GNSS) measurements for localization, navigation, and motion planning~\cite{national2021early, wu2020spoofing}. Civilian GNSS signals, however, are low-power and generally unauthenticated, making them vulnerable to spoofing attacks that can manipulate the perceived vehicle position and
mislead downstream navigation and control functions~\cite{abrar2024gps,dasgupta2022sensor}. Timely and reliable GNSS spoofing detection is therefore important for maintaining safe and trustworthy autonomous driving.

A central difficulty is that practical spoofing attacks may not manifest as isolated and immediately distinguishable anomalies~\cite{dasgupta2020prediction,clements2022carrier}. An attacker can introduce gradual or subtle deviations while preserving the apparent plausibility of individual GNSS observations. Although each observation may appear locally reasonable, the displacement implied by consecutive GNSS positions can become progressively inconsistent
with the motion measured by onboard sensors. The resulting attack evidence is therefore distributed across time and may overlap with residual variations caused by positioning noise, vehicle dynamics, and motion-sensor uncertainty. The main challenge is consequently not only to measure instantaneous GNSS--motion disagreement, but also to characterize how that disagreement evolves within a finite causal observation context.

Existing GNSS spoofing detectors have made important progress toward secure vehicle localization. GPS-IDS, for example, derives physical vehicle-behavior features and uses conventional machine-learning classifiers to distinguish normal and spoofed behavior in AV-GPS data~\cite{abrar2024gps}. Although physics-guided features improve interpretability, snapshot-based classification does not explicitly describe the temporal evolution of a
gradually changing GNSS--motion inconsistency. Sensor-fusion-based methods instead predict vehicle motion from in-vehicle sensors and compare the prediction with GNSS-derived motion. In~\cite{dasgupta2022sensor}, an LSTM-based motion estimate is used to obtain location discrepancies that are subsequently evaluated through a threshold-based decision mechanism. Such
methods exploit temporal information, but a single residual or prediction error does not explicitly separate the present inconsistency from its successive temporal variations. Moreover, fixed residual thresholds can be sensitive to normal changes in sensing and driving conditions. These limitations motivate a causal representation that preserves physically
interpretable residual evidence, models its different temporal orders separately, and subsequently learns their structured interaction.

To address this problem, we propose a causal high-order liquid evidence framework for GNSS spoofing detection. The framework first compares GNSS-implied displacement with onboard-motion-implied displacement and constructs an uncertainty-normalized GNSS--motion inconsistency residual. Rather than treating this residual as a single static feature, the method
forms zeroth-, first-, and second-order evidence describing the current residual level and its successive discrete temporal variations. The three orders are organized into separate evidence streams with order-dependent context selection and processed by separate adaptive liquid encoders within a fixed-length causal evidence window. The resulting order-specific temporal
states are then coupled hierarchically, allowing higher-order states to be progressively conditioned on lower-order context. Finally, the terminal fused representation yields the spoofing probability for the window endpoint.

The main contributions of this work are summarized as follows:

\begin{itemize}

    \item We formulate subtle GNSS spoofing detection as a causal displacement-consistency problem between GNSS and onboard motion, and derive physics-guided residual evidence from their disagreement.

    \item We construct a high-order evidence representation comprising the residual level and its first- and second-order discrete variations, with relevant contextual cues selected according to the evidence order.

    \item We design separate adaptive liquid encoders to model the within-window evolution of the three evidence orders, and hierarchically couple their hidden states for causal spoofing prediction at the window endpoint.

\end{itemize}

\section{Related Work }\label{sec:relatedwork}

\subsection{Static and Feature-Based GNSS Spoofing Detection}
\label{sec:static_gnss_detection}

GNSS spoofing detection is often formulated as a feature-based anomaly identification problem using signal-level, receiver-level, or vehicle-behavior information. Signal-based methods exploit indicators such as signal power, correlation distortion, direction of arrival, receiver-clock behavior, pseudorange consistency, and baseband correlator responses~\cite{schmidt2020gps,oligeri2022gps,radovs2024recent}. These approaches can be effective when spoofing produces identifiable signal-domain artifacts, but their applicability may depend on receiver access, observable signal characteristics, or assumptions about the attack configuration.

Learning-based detectors have also been developed using different feature sources. Some methods learn decision boundaries from GNSS signal or receiver measurements for spoofing detection and classification~\cite{shafique2021detecting,namagembe2026machine}, whereas others use tree-based models or learning-from-demonstration strategies to identify abnormal GPS behavior in vehicle-navigation scenarios~\cite{aissou2021tree,yang2023anomaly}. Infrastructure- and crowd-assisted approaches further compare GNSS measurements with trusted external information and apply anomaly-detection techniques such as Isolation Forest~\cite{wang2023infrastructure,oligeri2022gps}. A closely related example is GPS-IDS~\cite{abrar2024gps}, which combines physics-derived vehicle-behavior features with conventional machine-learning classifiers on AV-GPS data.

Overall, feature-based methods are effective when spoofing produces discriminative signal or vehicle-behavior descriptors. However, they do not explicitly characterize the multi-order temporal evolution of GNSS--motion residuals, which is important for detecting subtle and gradually developing attacks.

\subsection{Sensor-Fusion and Residual-Based Detection}
\label{sec:sensor_fusion_detection}

Sensor-fusion and residual-based methods detect GNSS spoofing by evaluating the consistency of GNSS measurements with independent motion or localization sources. In model-based approaches, pose or trajectory estimates are obtained from inertial sensors, vehicle dynamics, or multi-sensor localization pipelines, and deviations between the estimated state and the GNSS output are used as attack evidence~\cite{liu2019secure,narain2019security,clements2022carrier}. Related studies have also shown that GNSS/INS and multi-sensor fusion systems may remain vulnerable to carefully designed spoofing attacks that conform to vehicle-motion or fusion assumptions~\cite{shen2020drift,bauer2025effect}.

Learning-aided residual methods have been explored in autonomous-vehicle settings. Prediction-based approaches use in-vehicle measurements, including IMU, CAN, speed, steering, and accelerometer data, to estimate vehicle motion and compare it with GNSS-derived motion~\cite{dasgupta2020prediction,dasgupta2022sensor}. Other studies employ reinforcement learning or deep-learning-based localization-security models to improve detection under different attack conditions~\cite{dasgupta2022reinforcement,shabbir2023securing}.

These methods demonstrate the value of independent motion information for detecting GNSS inconsistencies. However, residuals or prediction errors are commonly used as direct decision variables, without explicitly separating the residual level from its first- and second-order temporal variations. This leaves room for a more structured representation of gradually evolving
GNSS--motion inconsistency.

\subsection{Temporal Modeling for Sequential Spoofing Detection}
\label{sec:temporal_spoofing_detection}

Temporal modeling has received increasing attention in spoofing and anomaly detection because attack evidence may develop across consecutive observations rather than appear at a single time step. In GNSS-related security, recurrent neural networks have been used to learn sequential patterns from receiver observations and low-cost spectrum measurements~\cite{calvo2020lstm}.
Motion-prediction-based LSTM models have also been employed to estimate vehicle displacement from GNSS, CAN, and IMU measurements and use the resulting discrepancy for spoofing detection~\cite{dasgupta2020prediction}. Related recurrent architectures have been applied to infer trajectory information from noisy mobile-sensor observations~\cite{jiang2022deeppose},
while more recent studies extend sequence modeling to satellite-characteristic data and drone-formation settings~\cite{dasgupta2023ai,wen2023lstm}.

Beyond GNSS security, temporal anomaly detection has progressed through recurrent, autoencoding, graph-based, and Transformer-based architectures. Stochastic recurrent networks support robust multivariate time-series anomaly detection~\cite{su2019robust}, while adversarial autoencoders enable unsupervised anomaly identification~\cite{audibert2020usad}. Graph-based temporal models capture dependencies among multiple variables~\cite{chen2021learning}, and Transformer-based approaches improve long-range dependency modeling and association-discrepancy analysis~\cite{xu2021anomaly,tuli2022tranad}.

These methods demonstrate the value of temporal dependency modeling for sequential anomaly detection. However, they generally operate on a unified feature sequence and learn a common temporal representation or anomaly score, rather than explicitly preserving the residual level and its successive discrete variations as separate temporal streams. They also do not directly model the ordered interaction among these residual-evidence streams.
\subsection{High-Order Temporal Modeling and Liquid Neural Networks}
\label{sec:high_order_liquid_modeling}

High-order and multi-scale temporal models have been developed to represent complex time-series variations beyond immediate step-to-step dependencies~\cite{he2019temporal,wu2022timesnet}. Continuous-time neural models describe sequence evolution through latent dynamical systems and can accommodate irregularly sampled observations~\cite{chen2018neural,rubanova2019latent}.
Liquid neural networks further introduce adaptive temporal dynamics through input-dependent time constants or closed-form state updates~\cite{hasani2021liquid,hasani2022closed}. These studies provide general mechanisms for flexible temporal representation, but they are not designed around physics-guided GNSS--motion evidence. In particular, they do not jointly preserve the residual level and its first- and second-order discrete variations through separate liquid encoders, followed by hierarchical interaction among their temporal states.

\section{Methods}\label{sec:Methodology}

\subsection{Problem Formulation and Causal Evidence-Window Setting}
\label{sec:problem_formulation}

Consider a ground vehicle observed at discrete time steps $t=1,\ldots,T$. At each time step, the vehicle provides a GNSS-side observation $g_t\in\mathbb{R}^{d_g}$ and an onboard motion observation $u_t\in\mathbb{R}^{d_u}$. The GNSS observation contains planar position
and signal-quality information, whereas $u_t$ contains vehicle-motion variables, such as velocity, yaw, yaw rate, and steering angle. We adopt a sensor-consistency threat model in which the attacker manipulates the GNSS-side observations, while the onboard motion measurements are not directly controlled, although they may remain noisy during normal operation.

Let $p_t^{g}\in\mathbb{R}^{2}$ denote the planar GNSS position extracted from $g_t$. Under normal conditions, the displacement obtained from consecutive GNSS positions should be kinematically consistent with that inferred from the onboard motion measurements. A gradual spoofing attack may preserve the apparent plausibility of individual GNSS observations
while progressively disrupting this consistency. Detection should therefore account for both the instantaneous GNSS--motion inconsistency and its temporal evolution.

Let $s(t)$ denote the first index of the continuous temporal segment containing time $t$. A causal evidence-construction operator $\mathcal{E}(\cdot)$ produces
\begin{equation}
\mathbf{e}_t = \mathcal{E}\!\left(
\{(g_{\tau},u_{\tau})\}_{\tau=s(t)}^{t} \right)
\in\mathbb{R}^{d_e},
\label{eq:causal_evidence}
\end{equation}
where $\mathbf{e}_t$ is the complete evidence vector at time $t$. As defined in Section~\ref{sec:high_order_evidence}, it contains the residual level, its first- and second-order temporal variations, and their associated physical and sensing cues, from which the three order-specific encoder inputs are selected. Every component of $\mathbf{e}_t$ is constructed using only the current and preceding observations within the same temporal segment.

For a prediction endpoint $t$, the latest $L$ complete evidence vectors are stacked as
\begin{equation}
\mathcal{X}_{t}= \left[
\begin{smallmatrix}
\mathbf{e}_{t-L+1}^{\top}\\[-1pt]
\vdots\\[-1pt]
\mathbf{e}_{t}^{\top}
\end{smallmatrix}
\right]
\in\mathbb{R}^{L\times d_e},\;
t-L+1\geq s(t),\qquad
\hat{p}_{t}=F_{\theta}(\mathcal{X}_{t}).
\label{eq:causal_detection}
\end{equation}
where $F_{\theta}(\cdot)$ denotes the proposed high-order liquid evidence model, $\hat{p}_{t}\in[0,1]$ is the spoofing probability assigned to the window endpoint, and $y_t\in\{0,1\}$ is the corresponding ground-truth label, with $y_t=1$ indicating spoofing. Adjacent windows may overlap within a segment, but no window crosses a segment boundary. Each window is processed independently, and all order-specific encoder states are initialized to zero before the first position of that window is processed.

\subsection{Physics-Guided GNSS--Motion Residual Evidence}
\label{sec:residual_construction}

The raw observations are first transformed into a physically meaningful GNSS--motion inconsistency signal. A gradual spoofing attack may keep individual GNSS positions apparently plausible while progressively making them inconsistent with the vehicle motion measured onboard. We therefore compare the displacement obtained from consecutive GNSS positions with that inferred from the onboard motion observations.

For $t>s(t)$, the GNSS-implied displacement is
\begin{equation}
\Delta p_t^{g}
=
p_t^{g}-p_{t-1}^{g},
\qquad
\Delta p_t^{g}\in\mathbb{R}^{2}.
\label{eq:gnss_displacement}
\end{equation}
Let $v_t^{u}\in\mathbb{R}^{2}$ denote the onboard velocity expressed in the same planar frame as $p_t^{g}$. It is obtained directly from the planar velocity components when available, $v_t^{u}=[v_{x,t},v_{y,t}]^{\top}$; otherwise, $v_t^{u}=s_t^{u}[\cos\psi_t^{u},\sin\psi_t^{u}]^{\top}$, where $s_t^{u}$ denotes the vehicle speed and $\psi_t^{u}$ denotes a causal heading estimate obtained from the onboard motion observations. The onboard-motion displacement over the same sampling interval is
\begin{equation}
\Delta p_t^{u}
=
\Delta t\,v_t^{u},
\qquad
\Delta p_t^{u}\in\mathbb{R}^{2},
\label{eq:motion_displacement}
\end{equation}
where $\Delta t$ denotes the sampling interval. At the first observation of each temporal segment, both displacement vectors are set to zero so that no information is propagated across segment boundaries.

The GNSS--motion inconsistency residual is defined as
\begin{equation}
r_t
=
\Delta p_t^{g}-\Delta p_t^{u},
\qquad
r_t\in\mathbb{R}^{2}.
\label{eq:raw_residual}
\end{equation}
Unlike the absolute GNSS position, $r_t$ directly measures the disagreement between GNSS-implied and onboard-motion-implied displacements. Under normal operation, the residual may vary because of positioning noise, vehicle dynamics, and motion-sensor uncertainty. During spoofing, its magnitude and temporal pattern may change as the received GNSS trajectory departs from the
vehicle motion.

Because the expected residual variability may differ between the two planar directions and across sensing conditions, we use the uncertainty-normalized residual
\begin{equation}
\eta_t
=
\left(
\Sigma_t+\epsilon I_2
\right)^{-1/2}r_t,
\qquad
\eta_t\in\mathbb{R}^{2},
\label{eq:normalized_residual}
\end{equation}
where $\Sigma_t\in\mathbb{R}^{2\times2}$ is a positive-definite residual uncertainty matrix and $\epsilon>0$ ensures numerical stability. Its nominal statistics are estimated from designated normal reference data and may be causally adjusted using available GNSS-quality information. Thus, $\eta_t$ expresses the displacement mismatch relative to its expected sensing variability rather than only in raw distance units.

The normalized residual $\eta_t$ provides the instantaneous physics-guided evidence for the subsequent high-order temporal representation.

\subsection{Causal High-Order Residual-Evidence Representation}
\label{sec:high_order_evidence}

The normalized residual $\eta_t$ measures the instantaneous GNSS--motion inconsistency. Because subtle spoofing may be more evident from how this inconsistency changes over time, we represent the residual level and its first- and second-order discrete variations as
\begin{equation}
\begin{aligned}
\eta_t^{(0)} &= \eta_t,\\
\eta_t^{(1)} &= \eta_t-\eta_{t-1},\\
\eta_t^{(2)} &= \eta_t^{(1)}-\eta_{t-1}^{(1)},
\end{aligned}
\qquad
\eta_t^{(k)}\in\mathbb{R}^{2},
\quad k\in\{0,1,2\}.
\label{eq:high_order_evidence}
\end{equation}
Here, $\eta_t^{(0)}$ describes the current mismatch, $\eta_t^{(1)}$ its step-to-step change, and $\eta_t^{(2)}$ the change in that variation. These finite differences are computed only within the current temporal segment, and unavailable initial differences are set to zero.

Let $\chi_t\in\mathbb{R}^{d_{\chi}}$ collect distinct causal contextual cues, including compact residual, uncertainty, displacement, and sensing-quality information not already represented by the three residual-evidence orders and their magnitudes. The complete evidence vector introduced in \eqref{eq:causal_evidence} is
\begin{equation}
\begin{aligned}
\mathbf{e}_t
={}&
\left[
(\eta_t^{(0)})^{\top},
\|\eta_t^{(0)}\|_2,
(\eta_t^{(1)})^{\top},
\|\eta_t^{(1)}\|_2,
\right.\\[-2pt]
&\left.
(\eta_t^{(2)})^{\top},
\|\eta_t^{(2)}\|_2,
\chi_t^{\top}
\right]^{\top}
\in\mathbb{R}^{d_e}.
\end{aligned}
\label{eq:complete_evidence_vector}
\end{equation}
Each distinct quantity is included once in $\mathbf e_t$.

The encoder inputs are constructed according to the role of each evidence order:
\begin{equation}
\begin{aligned}
\xi_t^{(0)}
&=
\left[
(\eta_t^{(0)})^{\top},
\|\eta_t^{(0)}\|_2,
(\chi_t^{(0)})^{\top}
\right]^{\top},\\
\xi_t^{(1)}
&=
\left[
(\eta_t^{(1)})^{\top},
\|\eta_t^{(1)}\|_2,
(\chi_t^{(1)})^{\top}
\right]^{\top},\\
\xi_t^{(2)}
&=
\left[
(\eta_t^{(2)})^{\top},
\|\eta_t^{(2)}\|_2
\right]^{\top},
\end{aligned}
\qquad
\xi_t^{(k)}\in\mathbb{R}^{d_k}.
\label{eq:order_specific_inputs}
\end{equation}
Here, $\chi_t^{(0)}$ and $\chi_t^{(1)}$ are fixed subvectors of $\chi_t$ associated with the residual level and first-order variation, respectively. The second-order stream remains focused on the highest-order variation without additional contextual cues. A contextual
cue may support more than one stream while appearing only once in $\mathbf e_t$.

For a causal window ending at time $t$, as defined in \eqref{eq:causal_detection}, the corresponding order-specific sequence is
\begin{equation}
\mathcal{X}_t^{(k)}
=
\begin{bmatrix}
(\xi_{t-L+1}^{(k)})^{\top}\\
\vdots\\
(\xi_t^{(k)})^{\top}
\end{bmatrix}
\in\mathbb{R}^{L\times d_k},
\qquad
k\in\{0,1,2\}.
\label{eq:order_specific_windows}
\end{equation}
Thus, the three sequences are order-specific views of the same causal evidence window and are processed by their corresponding temporal encoders.

\subsection{Order-Specific Adaptive Liquid Temporal Encoding}
\label{sec:liquid_temporal_encoding}

The zeroth-, first-, and second-order evidence streams describe different aspects of residual evolution and may exhibit different temporal persistence. The residual level may remain elevated across several observations, whereas its first- and second-order variations may change
more rapidly. Encoding all three orders through a single recurrent state could obscure these distinct behaviors. We therefore assign an independent adaptive liquid encoder to each order-specific sequence $\mathcal{X}_t^{(k)}$, $k\in\{0,1,2\}$, allowing each evidence order to learn its own temporal response before cross-order coupling.

For a window ending at time $t$, let
\[
\xi_{t,\ell}^{(k)}
=
\xi_{t-L+\ell}^{(k)},
\qquad
\ell=1,\ldots,L,
\quad
k\in\{0,1,2\},
\]
denote the $\ell$th input of the $k$th evidence sequence, and let $h_{t,\ell}^{(k)}\in\mathbb{R}^{d_h}$ denote its hidden state. Consistent with the independent-window setting in Section~\ref{sec:problem_formulation}, each encoder is initialized as $h_{t,0}^{(k)}=\mathbf{0}_{d_h}$.

At each window position, an evidence- and state-dependent time constant and its corresponding update factor are computed as
\begin{equation}
\begin{aligned}
\tau_{t,\ell}^{(k)}
&=
\tau_{\min}\mathbf{1}_{d_h}
+
\operatorname{softplus}\!\left(
A_{\xi}^{(k)}\xi_{t,\ell}^{(k)}
+
A_{h}^{(k)}h_{t,\ell-1}^{(k)}
+
b_{\tau}^{(k)}
\right),\\
\lambda_{t,\ell}^{(k)}
&=
\mathbf{1}_{d_h}
-
\exp\!\left(
-\delta_{\mathrm{L}}\oslash\tau_{t,\ell}^{(k)}
\right),
\end{aligned}
\label{eq:adaptive_liquid_factor}
\end{equation}
where $\tau_{\min}>0$ provides a positive lower bound on the time constant, $\delta_{\mathrm{L}}>0$ is the discretization step of the liquid update, and $\oslash$ denotes element-wise division. The softplus and exponential operations are applied element-wise, giving $\lambda_{t,\ell}^{(k)}\in(0,1)^{d_h}$.

The order-specific liquid state is updated by
\begin{equation}
\begin{aligned}
h_{t,\ell}^{(k)}
={}&
\left(
\mathbf{1}_{d_h}-\lambda_{t,\ell}^{(k)}
\right)
\odot h_{t,\ell-1}^{(k)}\\
&+
\lambda_{t,\ell}^{(k)}
\odot
\tanh\!\left(
W_{\xi}^{(k)}\xi_{t,\ell}^{(k)}
+
W_{h}^{(k)}h_{t,\ell-1}^{(k)}
+
b_{h}^{(k)}
\right),
\end{aligned}
\label{eq:liquid_state_update}
\end{equation}
where $\odot$ denotes element-wise multiplication. A larger time constant produces a smaller update factor and therefore preserves more information from the preceding state, whereas a smaller time constant allows the encoder to respond more strongly to the current evidence. Because the time constant depends on both the current input and the preceding state, the effective memory adapts across positions within the evidence window.

All parameters carrying the superscript $k$ are learned independently for the corresponding evidence order. Consequently, $h_{t,\ell}^{(0)}$, $h_{t,\ell}^{(1)}$, and $h_{t,\ell}^{(2)}$ separately summarize the within-window evolution of the residual level and its first- and second-order variations. 

\subsection{Hierarchical Cross-Order Temporal-State Coupling}
\label{sec:cross_order_coupling}

The three liquid encoders produce complementary temporal states describing the residual level and its first- and second-order variations. Although these hidden states are generated by separate encoders before coupling and are not themselves finite differences of one another, their underlying evidence streams follow an ordered construction: the first-order evidence
is derived from the zeroth-order evidence, and the second-order evidence is derived from the first-order variation. We therefore introduce a hierarchical coupling mechanism as an inductive bias that progressively conditions higher-order temporal states on lower-order context. Unlike
direct fusion of the uncoupled states, this design explicitly reflects the order in which the residual-evidence streams are constructed.

At position $\ell$ of the window ending at time $t$, the zeroth-order state provides the base context, $\bar{h}_{t,\ell}^{(0)}=h_{t,\ell}^{(0)}$. The remaining coupled states are obtained as
\begin{equation}
\begin{aligned}
\bar{h}_{t,\ell}^{(1)}
&=
\tanh\!\left(
W_{c}^{(1)}
\begin{bmatrix}
h_{t,\ell}^{(1)}\\
\bar{h}_{t,\ell}^{(0)}
\end{bmatrix}
+
b_{c}^{(1)}
\right),\\
\bar{h}_{t,\ell}^{(2)}
&=
\tanh\!\left(
W_{c}^{(2)}
\begin{bmatrix}
h_{t,\ell}^{(2)}\\
\bar{h}_{t,\ell}^{(1)}
\end{bmatrix}
+
b_{c}^{(2)}
\right),
\end{aligned}
\label{eq:hierarchical_cross_order_coupling}
\end{equation}
where $\bar{h}_{t,\ell}^{(k)}\in\mathbb{R}^{d_h}$ and $W_c^{(1)}$, $W_c^{(2)}$, $b_c^{(1)}$, and $b_c^{(2)}$ are learnable coupling parameters. Thus, the first-order state is
interpreted in the context of the current residual level, while the second-order state is conditioned on the representation that already combines the lower two evidence orders.

The three coupled states are fused into an order-aware representation:
\begin{equation}
f_{t,\ell}
=
\tanh\!\left(
W_f
\begin{bmatrix}
\bar{h}_{t,\ell}^{(0)}\\
\bar{h}_{t,\ell}^{(1)}\\
\bar{h}_{t,\ell}^{(2)}
\end{bmatrix}
+
b_f
\right),
\qquad
f_{t,\ell}\in\mathbb{R}^{d_f},
\label{eq:fused_cross_order_state}
\end{equation}
where $W_f$ and $b_f$ are learnable fusion parameters. Applying the same coupling at every window position yields the fused sequence $\mathcal{F}_t=[f_{t,1},\ldots,f_{t,L}]^{\top} \in\mathbb{R}^{L\times d_f}$. The sequence $\mathcal{F}_t$ therefore preserves the within window temporal organization while integrating the residual level and its successive variations at each position.

\subsection{Endpoint-Level Spoofing Prediction and Learning Objective}
\label{sec:detection_objective}

The hierarchical coupling stage produces the fused temporal sequence $\mathcal{F}_t$ for the causal evidence window ending at time $t$. Because each window is assigned the ground-truth label of its endpoint, the terminal fused state $f_{t,L}$ is used to estimate the spoofing
probability at that endpoint. The preceding positions receive no direct position-wise supervision, but they influence $f_{t,L}$ through the order-specific recurrent dynamics and therefore contribute to the endpoint loss during training.

Let $\rho_{\psi}:\mathbb{R}^{d_f}\rightarrow\mathbb{R}$ denote the parameterized decision map applied to the terminal fused representation. The endpoint logit and spoofing probability are defined as
\begin{equation}
o_t
=
\rho_{\psi}\!\left(f_{t,L}\right),
\qquad
\hat{p}_t
=
\sigma(o_t)
=
F_{\theta}(\mathcal{X}_t),
\label{eq:endpoint_spoofing_probability}
\end{equation}
where $\psi$ denotes the decision-map parameters and is included in the complete trainable parameter set $\theta$. The resulting $\hat{p}_t\in[0,1]$ is the spoofing probability assigned to endpoint $t$.

Let $\mathcal{T}_{\mathrm{tr}}$ denote the set of endpoint indices of the training windows. To account for class imbalance, the complete detector is learned using the positive-class-weighted binary cross-entropy objective
\begin{equation}
\mathcal{L}(\theta)
=
-\frac{1}{|\mathcal{T}_{\mathrm{tr}}|}
\sum_{t\in\mathcal{T}_{\mathrm{tr}}}
\left[
\omega_{+}y_t\log\hat{p}_t
+
(1-y_t)\log(1-\hat{p}_t)
\right],
\label{eq:weighted_bce_loss}
\end{equation}
where $\omega_{+}=N_{0}/N_{1}$, and $N_{0}$ and $N_{1}$ denote the numbers of normal and spoofed training windows, respectively. Although supervision is imposed only on the endpoint prediction, all learnable parameters of the order-specific liquid encoders, hierarchical coupling mechanism, fusion layer, and decision map are optimized jointly.

During inference, the binary endpoint decision is $\widehat{y}_t=\mathbb{I}(\hat{p}_t\geq\gamma)$, where $\mathbb{I}(\cdot)$ is the indicator function and $\gamma$ is a pre-specified operating threshold. The threshold is fixed before held-out evaluation and applied unchanged to all evaluation subsets. Because $\mathcal{F}_t$ is constructed solely from $\mathcal{X}_t$, the prediction uses no observation after endpoint $t$. The detector therefore produces causal endpoint decisions from a finite temporal context of length $L$, without propagating hidden states between adjacent prediction windows.

\begin{figure*}[!t]
\centering
\includegraphics[width=1.0\textwidth]{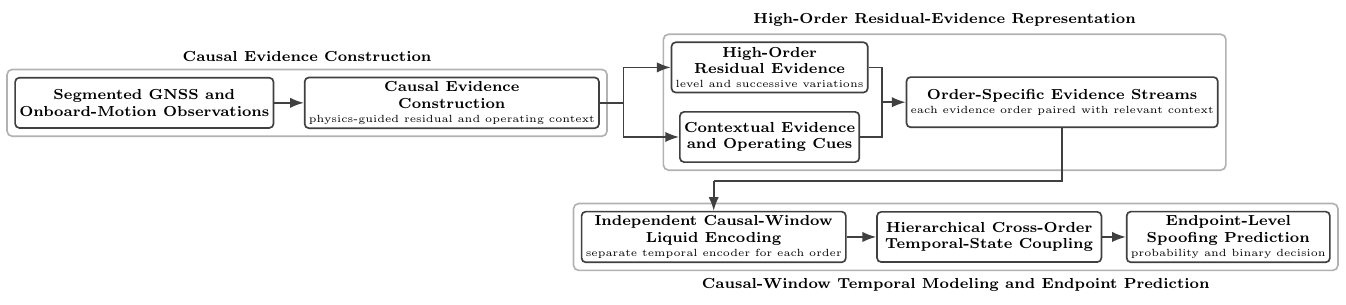}
\caption{Overview of the proposed causal high-order liquid evidence framework for GNSS spoofing detection.}
\label{fig:method_overview}
\end{figure*}

Fig.~\ref{fig:method_overview} illustrates the complete workflow of the proposed framework. The overall inference procedure of the proposed framework is summarized in Algorithm~\ref{alg:causal_high_order_liquid_detection}. 

\begin{algorithm}[t]
\caption{Causal High-Order Liquid Evidence Detection}
\label{alg:causal_high_order_liquid_detection}
\small
\begin{algorithmic}[1]

\Require Segmented observations $\{(g_t,u_t)\}_{t=1}^{T}$;
window length $L$; sampling interval $\Delta t$; liquid-update step
$\delta_{\mathrm{L}}$; residual-uncertainty parameters defining $\Sigma_t$; regularization constant $\epsilon$; trained detector
parameters $\theta$; threshold $\gamma$

\Ensure Endpoint probabilities and decisions
$\{(\hat{p}_t,\widehat{y}_t)\}_{t\in\mathcal{T}_L}$

\Statex \textbf{Causal evidence construction}

\For{each continuous temporal segment}

    \State Compute $\Delta p_t^{g}$ and $\Delta p_t^{u}$ for all indices
    within the segment using Eqs.~\eqref{eq:gnss_displacement} and
    \eqref{eq:motion_displacement}

    \State Construct residual $r_t$ and its uncertainty-normalized form $\eta_t$
    using Eqs.~\eqref{eq:raw_residual} and
    \eqref{eq:normalized_residual}

    \State Derive high-order residual evidence
    $\{\eta_t^{(0)},\eta_t^{(1)},\eta_t^{(2)}\}$ using
    Eq.~\eqref{eq:high_order_evidence}

    \State Form the context $\chi_t$, complete evidence vector
    $\mathbf{e}_t$, and order-specific inputs
    $\{\xi_t^{(k)}\}_{k=0}^{2}$ using
    Eqs.~\eqref{eq:complete_evidence_vector} and
    \eqref{eq:order_specific_inputs}

\EndFor

\State Define endpoint set
$\mathcal{T}_L
=
\{t\in\{1,\ldots,T\}\mid t-L+1\geq s(t)\}$

\Statex \textbf{Independent causal-window detection}

\For{each endpoint $t\in\mathcal{T}_L$}

    \State Initialize
    $h_{t,0}^{(k)}\gets\mathbf{0}_{d_h}$ for
    $k\in\{0,1,2\}$

    \For{$\ell=1$ to $L$}

        \State Set
        $\xi_{t,\ell}^{(k)}
        \gets\xi_{t-L+\ell}^{(k)}$
        for $k\in\{0,1,2\}$

        \For{$k=0$ to $2$}

            \State Compute
            $\tau_{t,\ell}^{(k)}$ and
            $\lambda_{t,\ell}^{(k)}$ using
            Eq.~\eqref{eq:adaptive_liquid_factor}

            \State Update $h_{t,\ell}^{(k)}$ using
            Eq.~\eqref{eq:liquid_state_update}

        \EndFor

        \State Compute
        $\{\bar{h}_{t,\ell}^{(k)}\}_{k=0}^{2}$ using
        Eq.~\eqref{eq:hierarchical_cross_order_coupling}

        \State Obtain the fused state $f_{t,\ell}$ using
        Eq.~\eqref{eq:fused_cross_order_state}

    \EndFor

    \State Compute
    $\hat{p}_t
    =
    \sigma\!\left(\rho_{\psi}(f_{t,L})\right)$
    using Eq.~\eqref{eq:endpoint_spoofing_probability}

    \State Set
    $\widehat{y}_t
    \gets
    \mathbb{I}\!\left(\hat{p}_t\geq\gamma\right)$

\EndFor

\State \Return
$\{(\hat{p}_t,\widehat{y}_t)\}_{t\in\mathcal{T}_L}$

\end{algorithmic}
\end{algorithm}

\section{Experimental Results and Discussion}

\subsection{Dataset, Evaluation Protocol, and Metrics}
\label{sec:dataset_protocol}

We evaluate the proposed method on the three subsets of the AV-GPS-Dataset released with GPS-IDS~\cite{abrar2024gps}. The dataset contains synchronized GNSS and onboard vehicle-motion measurements collected under normal driving and practical GPS spoofing conditions. Because the
subsets differ in size, recording structure, and acquisition conditions, each is assigned a distinct experimental role, as summarized in Table~\ref{tab:dataset_protocol}.

\begin{table}[H]
\centering
\caption{Raw statistics and designated evaluation roles of the
AV-GPS datasets~\cite{abrar2024gps}.}
\label{tab:dataset_protocol}
\footnotesize
\setlength{\tabcolsep}{3.7pt}
\begin{tabular}{lrrrp{2.35cm}}
\toprule
\textbf{Dataset} &
\textbf{Total} &
\textbf{Normal} &
\textbf{Attack} &
\textbf{Evaluation Role} \\
\midrule
Dataset 1 &
62,042 &
46,287 &
15,755 &
Development and controlled in-domain test \\

Dataset 2 &
6,890 &
5,184 &
1,706 &
Cross-location transfer stress test \\

Dataset 3 &
636 &
231 &
405 &
Transition-aware sequential case study \\
\bottomrule
\end{tabular}
\end{table}

Residual and high-order evidence are constructed causally within each continuous temporal segment. A length-\(L\) evidence window is retained only when all of its positions belong to the same segment. Each valid window is processed independently and assigned the ground-truth label of its endpoint; no evidence difference, window, or encoder state crosses a segment boundary.

Dataset~1 is the largest and most heterogeneous subset. The observations are separated at recording discontinuities and label boundaries, yielding 29 non-overlapping temporal segments that are assigned disjointly to the training, validation, and test partitions. The resulting allocation contains 21 training segments, 5 validation segments, and 3 held-out test segments.
With \(L=10\), these partitions produce 43,462 training windows, 7,510 validation windows, and 10,813 held-out test windows, respectively. All data-derived preprocessing quantities are estimated exclusively from the Dataset~1 training partition, and the model checkpoint is selected using the validation-set F1-score. The Dataset~1 test partition is used as a controlled
in-domain benchmark and is not interpreted as evidence of cross-location generalization.

Dataset~2 was collected at a different location under changed driving and GNSS-reception conditions, including partial tree obstruction ~\cite{abrar2024gps}. Its normal and spoofed observations originate from separate recording sessions rather than continuous normal-to-attack
transitions. Dataset~2 is therefore used exclusively as a cross-location stress test and is not involved in training, preprocessing estimation, checkpoint selection, threshold adjustment, or attack-onset analysis. The preprocessing quantities and selected checkpoint obtained from Dataset~1 are applied unchanged to its 6,838 valid causal windows.

Dataset~3 contains one continuous driving sequence with two labeled normal-to-spoofing transitions. It is excluded from training, preprocessing estimation, checkpoint selection, and threshold determination, and is used only for transition-aware sequential analysis. With \(L=10\), the sequence produces 627 valid causal windows. Its temporal structure supports analysis of prediction evolution and response delay around attack onset. Because only two attack transitions are available, the delay results are interpreted as illustrative event-level evidence rather than as a statistically comprehensive early-detection benchmark.

A common pre-specified decision threshold of \(\gamma=0.5\) is used for the proposed method and the controlled temporal baselines. Performance is reported using accuracy, precision, recall, F1-score, false-alarm rate (FAR), and false-negative rate (FNR), with F1-score, FAR, and FNR emphasized in the main comparisons. ROC-AUC and PR-AUC are additionally reported for Dataset~2 to separate threshold-independent score discrimination from performance at the
fixed operating threshold. For Dataset~3, detection delay is measured from each labeled attack onset to the first endpoint at or after that onset whose spoofing probability reaches the fixed threshold.

\subsection{Comparison with Temporal Baselines}
\label{sec:overall_detection}

Because the proposed detector models the temporal evolution of GNSS--motion inconsistency, we compare it with LSTM, GRU, TCN, and Transformer baselines, representing recurrent, temporal-convolutional, and attention-based sequence-modeling families. This comparison examines whether
the proposed framework offers benefits beyond generic temporal encoding. Table~\ref{tab:overall_results} presents the main results on Dataset~1 and Dataset~3, while cross-location transfer on Dataset~2 is analyzed separately.

\begin{table}[H]
\centering
\caption{Detection performance comparison on Dataset~1 and Dataset~3.}
\label{tab:overall_results}
\footnotesize
\setlength{\tabcolsep}{4.7pt}
\begin{tabular}{llccccc}
\toprule
\textbf{Data} & \textbf{Model} & \textbf{F1}$\uparrow$ &
\textbf{FAR}$\downarrow$ & \textbf{FNR}$\downarrow$ &
\textbf{Acc.}$\uparrow$ & \textbf{Prec.}$\uparrow$ \\
\midrule

\multirow{5}{*}{Dataset 1}
& Proposed    & \textbf{0.9535} & \textbf{0.0086} &
\textbf{0.0432} & \textbf{0.9864} & \textbf{0.9502} \\
& LSTM        & 0.9415 & 0.0107 & 0.0546 & 0.9829 & 0.9377 \\
& GRU         & 0.9401 & \textit{0.0104} & 0.0590 & 0.9825 & 0.9392 \\
& TCN         & \textit{0.9478} & 0.0105 & \textit{0.0438} &
\textit{0.9846} & 0.9395 \\
& Transformer & 0.9437 & \textit{0.0104} & 0.0521 & 0.9835 &
\textit{0.9396} \\
\midrule

\multirow{5}{*}{Dataset 3}
& Proposed    & \textbf{0.9777} & \textbf{0.0315} &
\textit{0.0272} & \textbf{0.9713} & \textbf{0.9825} \\
& LSTM        & 0.9692 & 0.0586 & 0.0296 & 0.9601 & 0.9680 \\
& GRU         & 0.9727 & \textit{0.0405} & 0.0321 &
\textit{0.9649} & \textit{0.9776} \\
& TCN         & \textit{0.9731} & 0.0676 & \textbf{0.0173} &
\textit{0.9649} & 0.9637 \\
& Transformer & 0.9391 & 0.0991 & 0.0667 & 0.9219 & 0.9450 \\
\bottomrule
\end{tabular}
\end{table}

On Dataset~1, the proposed method is consistently strongest across the reported metrics. The simultaneous improvement in F1-score, FAR, and FNR indicates that the gain is not obtained by reducing one error type at the expense of the other. This supports the effectiveness of the integrated high-order liquid architecture relative to generic temporal models.

Dataset~3 reveals a more explicit detection trade-off. TCN attains the lowest FNR, but its higher FAR and lower precision indicate that the increased attack sensitivity is accompanied by more false alarms. The proposed method provides the strongest overall balance across F1-score, FAR, accuracy, and precision, while retaining the second-lowest FNR. Because Dataset~3 contains only two attack transitions, these findings are interpreted as sequential case-study evidence rather than as a broad generalization result.

\subsection{Cross-Location Stress Test on Dataset~2}
\label{sec:dataset2_stress}

Dataset~2 examines whether detectors developed on Dataset~1 retain their discriminative behavior under changed acquisition conditions. Following the protocol defined above, all models are transferred without location-specific retraining, model reselection, or parameter adjustment. This experiment therefore evaluates whether attack discrimination and false-alarm control transfer consistently to a different operating environment. The resulting cross location performance is reported in Table~\ref{tab:dataset2_stress}.

\begin{table}[H]
\centering
\caption{Cross-location transfer performance on Dataset~2.}
\label{tab:dataset2_stress}
\footnotesize
\setlength{\tabcolsep}{3.6pt}
\resizebox{\columnwidth}{!}{%
\begin{tabular}{lccccccc}
\toprule
\textbf{Model} &
\textbf{F1}$\uparrow$ &
\textbf{FAR}$\downarrow$ &
\textbf{FNR}$\downarrow$ &
\textbf{Acc.}$\uparrow$ &
\textbf{Prec.}$\uparrow$ &
\textbf{ROC-AUC}$\uparrow$ &
\textbf{PR-AUC}$\uparrow$ \\
\midrule
Proposed    & 0.8368 & 0.1260 & 0.0006 & 0.9047 & 0.7196 & 0.99987 & 0.99970 \\
LSTM        & 0.8594 & 0.1051 & 0.0018 & 0.9202 & 0.7545 & 0.99957 & 0.99898 \\
GRU         & 0.8311 & 0.1312 & 0.0006 & 0.9007 & 0.7114 & 0.99978 & 0.99949 \\
TCN         & 0.8775 & 0.0904 & 0.0000 & 0.9317 & 0.7817 & 0.99996 & 0.99987 \\
Transformer & 0.8762 & 0.0910 & 0.0012 & 0.9310 & 0.7804 & 0.99966 & 0.99917 \\
\bottomrule
\end{tabular}%
}
\end{table}

All models retain high attack sensitivity on Dataset~2, with FNR values below 0.002. The main performance differences arise from normal observations: TCN and Transformer produce fewer false alarms and consequently achieve higher F1-scores and precision. The proposed method retains near-complete attack detection and obtains the second-highest ROC-AUC and PR-AUC, but its higher FAR shows weaker transfer of false-alarm control.

The near-ceiling ROC-AUC and PR-AUC values show that normal and spoofed windows remain strongly separable across locations. For the proposed method, the contrast between strong score discrimination and weaker F1, precision, and FAR suggests that the main transfer degradation is
associated with responses to normal observations rather than with a loss of attack separability. Dataset~2 therefore reveals false-alarm robustness under changed reception conditions as a remaining limitation of the proposed framework.

\subsection{Role of High-Order Residual Evidence}
\label{sec:ablation_components}

To isolate the contribution of the proposed high-order residual-evidence representation, we construct three nested input variants while keeping all non-ablated model components and training settings unchanged. The \emph{Auxiliary Evidence Only} variant removes $\eta_t^{(k)}$ and $\|\eta_t^{(k)}\|_2$ for all $k\in\{0,1,2\}$, retaining only the contextual evidence selected from $\chi_t$. The \emph{Residual Level Only} variant additionally includes
$\eta_t^{(0)}$ and $\|\eta_t^{(0)}\|_2$. The \emph{Level + First Order} variant includes the zeroth- and first-order residual components but excludes $\eta_t^{(2)}$ and $\|\eta_t^{(2)}\|_2$. The complete model includes all three residual-order vectors and their corresponding magnitudes. These comparisons evaluate the incremental contributions of the residual level and its first- and second-order discrete variations beyond the
contextual evidence. Table~\ref{tab:ablation_results} reports the results on Dataset~1 and Dataset~3, while Dataset~2 remains reserved for whole-model cross-location evaluation.

\begin{table}[t]
\centering
\caption{Ablation of the high-order residual-evidence representation on
Dataset~1 and Dataset~3.}
\label{tab:ablation_results}
\footnotesize
\setlength{\tabcolsep}{2.5pt}
\begin{tabular}{llccccc}
\toprule
\textbf{Dataset} & \textbf{Variant} & \textbf{F1}$\uparrow$ &
\textbf{FAR}$\downarrow$ & \textbf{FNR}$\downarrow$ &
\textbf{Acc.}$\uparrow$ & \textbf{Prec.}$\uparrow$ \\
\midrule

\multirow{4}{*}{Dataset 1}
& Proposed
& \textbf{0.9535} & \textbf{0.0086} & \textbf{0.0432}
& \textbf{0.9864} & \textbf{0.9502} \\
& Auxiliary Evidence Only
& 0.9276 & 0.0092 & 0.0883 & 0.9793 & 0.9441 \\
& Residual Level Only
& 0.9350 & 0.0088 & 0.0768 & 0.9813 & 0.9472 \\
& Level + First Order
& 0.9272 & 0.0176 & 0.0463 & 0.9782 & 0.9021 \\
\midrule

\multirow{4}{*}{Dataset 3}
& Proposed
& \textbf{0.9777} & \textbf{0.0315} & 0.0272
& \textbf{0.9713} & \textbf{0.9825} \\
& Auxiliary Evidence Only
& 0.9653 & 0.0541 & 0.0395 & 0.9553 & 0.9701 \\
& Residual Level Only
& 0.9729 & 0.0541 & \textbf{0.0247} & 0.9649 & 0.9705 \\
& Level + First Order
& 0.9717 & 0.0586 & \textbf{0.0247} & 0.9633 & 0.9681 \\
\bottomrule
\end{tabular}
\end{table}

On Dataset~1, introducing the residual-level components reduces missed detections relative to the auxiliary-evidence-only variant. Adding the first-order components further improves attack sensitivity, but weakens false-alarm control and precision. Incorporating the second-order components produces the strongest overall performance, indicating that the three
residual orders provide complementary rather than uniformly additive information.

Dataset~3 exhibits a related trade-off. The variants excluding the second-order components attain slightly lower FNR, but their higher FAR and lower F1-score, accuracy, and precision reflect a less favorable overall detection balance. The second-order evidence therefore does not improve every individual metric; instead, it contributes to a more balanced
discrimination between normal and spoofed states. Collectively, these results support the integration of high-order residual evidence with the contextual evidence.

\subsection{Role of Adaptive Liquid Temporal Encoding}
\label{sec:temporal_evolution}

The comparison in Table~\ref{tab:overall_results} evaluates complete sequence models, whereas this experiment isolates the temporal encoder within the proposed framework. The high-order residual inputs, order-specific contextual evidence, three-stream organization,
hierarchical cross-order coupling, and detection head are retained; only the adaptive liquid encoder in each order-specific stream is replaced by an LSTM or GRU encoder. The replacements follow the same input construction, training, model-selection, and operating-threshold
protocol. This controlled comparison therefore examines whether adaptive liquid encoding contributes beyond the proposed evidence representation and cross-order fusion structure. Table~\ref{tab:temporal_evolution} reports the results on Dataset~1.

\begin{table}[H]
\centering
\caption{Controlled comparison of temporal encoders on
Dataset~1.}
\label{tab:temporal_evolution}
\footnotesize
\setlength{\tabcolsep}{3.6pt}
\begin{tabular}{lccccc}
\toprule
\textbf{Encoder} &
\textbf{F1}$\uparrow$ &
\textbf{FAR}$\downarrow$ &
\textbf{FNR}$\downarrow$ &
\textbf{Acc.}$\uparrow$ &
\textbf{Prec.}$\uparrow$ \\
\midrule
LSTM Replacement
& 0.9390 & 0.0132 & 0.0463 & 0.9820 & 0.9249 \\

GRU Replacement
& 0.9372 & 0.0118 & 0.0571 & 0.9816 & 0.9316 \\

\textbf{Adaptive Liquid Encoder}
& \textbf{0.9535} & \textbf{0.0086} & \textbf{0.0432}
& \textbf{0.9864} & \textbf{0.9502} \\
\bottomrule
\end{tabular}
\end{table}

Both recurrent replacements degrade every reported metric. In particular, their higher FAR and FNR show that the advantage of adaptive liquid encoding is not obtained by improving one error type at the expense of the other. Because the evidence representation and cross-order coupling are retained, the results support the temporal encoder as a complementary contributor to the complete framework. This finding is consistent with its intended role of modeling the distinct within-window evolution of the residual level and its first- and second-order variations.

\subsection{Sequential Transition Case Study on Dataset~3}
\label{sec:early_detection}

Aggregate metrics summarize overall detection errors but do not reveal when the detector responds relative to attack onset. We therefore examine the two labeled normal-to-spoofing transitions in the continuous Dataset~3 sequence. For each event, detection delay is measured from the labeled onset endpoint to the first endpoint at or after that onset whose spoofing probability reaches the decision threshold. Because successive causal windows advance by one observation, the delay is reported in endpoint steps, with each step corresponding to one sampling interval. The resulting event-level detection measurements are reported in Table~\ref{tab:early_detection}.

The first detected endpoints occur four and three sampling intervals after the respective labeled onsets. Thus, in both available events, the detector responds within four endpoint steps. Because Dataset~3 contains only two normal-to-spoofing transitions, these measurements are interpreted as illustrative event-level evidence rather than as a statistically general early-detection result.

\begin{table}[H]
\centering
\caption{Transition-level detection results on Dataset~3.}
\label{tab:early_detection}
\footnotesize
\setlength{\tabcolsep}{2.2pt}
\renewcommand{\arraystretch}{1.1}
\begin{tabular}{lcccc}
\toprule
\textbf{Event} &
\shortstack{\textbf{Onset}\\\textbf{Endpoint}} &
\shortstack{\textbf{Detected}\\\textbf{Endpoint}} &
\shortstack{\textbf{Delay}\\\textbf{(steps)}} &
\shortstack{\textbf{Probability}\\\textbf{at Detection}} \\
\midrule
T1 & 133 & 137 & 4 & 0.8511 \\
T2 & 403 & 406 & 3 & 0.8699 \\
\bottomrule
\end{tabular}
\end{table}

\begin{figure}[H]
\centering
\includegraphics[width=\columnwidth]{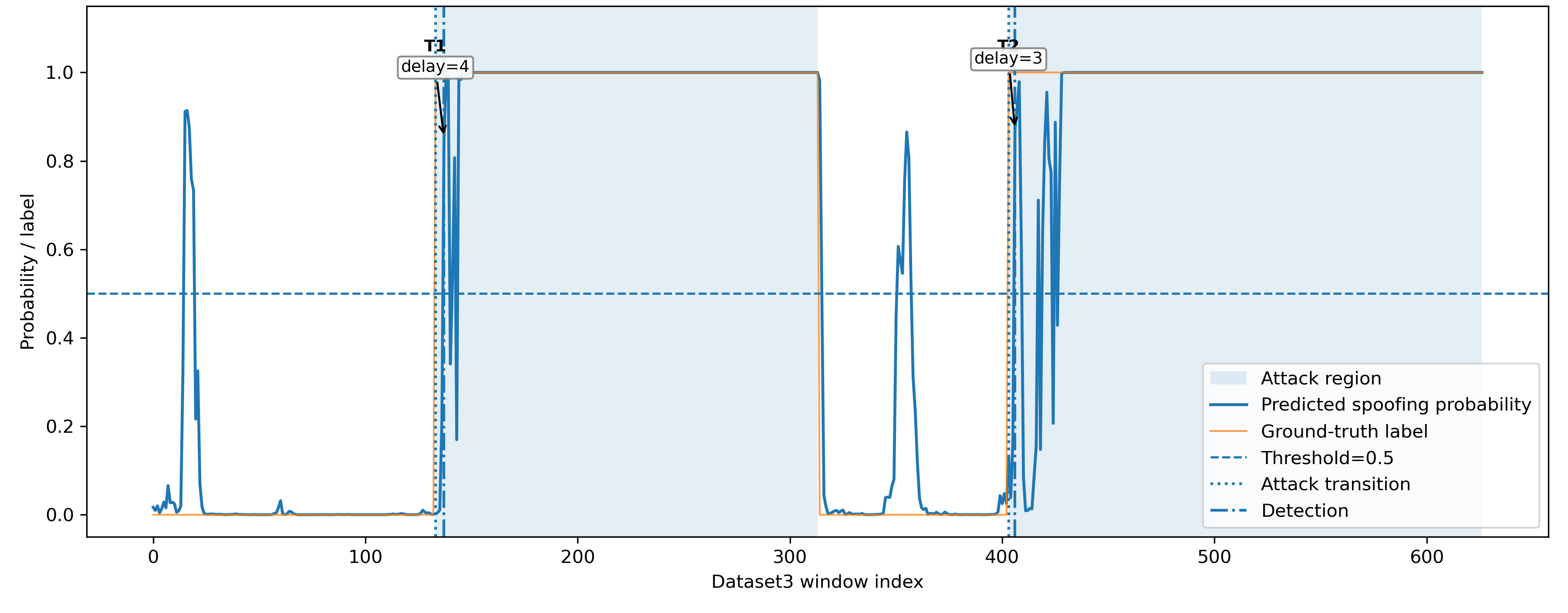}
\caption{Sequential detection response on Dataset~3. Shaded regions indicate spoofed intervals, dotted vertical lines mark the labeled attack onsets, and dash-dotted vertical lines mark the corresponding first detections. The horizontal dashed line denotes the decision threshold.}
\label{fig:early_detection}
\end{figure}

Figure~\ref{fig:early_detection} presents the spoofing-probability trajectory over the complete sequence. Following each labeled onset, the predicted probability crosses the threshold and remains predominantly elevated across the corresponding spoofed interval. Above-threshold excursions also occur during normal operation, consistent with the nonzero false-alarm rate reported for Dataset~3. The sequence-level analysis therefore illustrates both the detector's response near attack onset and its remaining sensitivity to transient variations under normal conditions.

\subsection{Sensitivity to Key Design Parameters}
\label{subsec:sensitivity}

We examine the sensitivity of the proposed detector to the causal window length \(L\), hidden dimension \(d_h\), and minimum liquid time constant \(\tau_{\min}\) using a one-factor-at-a-time protocol. In each experiment, only the parameter under study is varied, while all remaining settings are kept fixed. The results are summarized in Table~\ref{tab:sensitivity_analysis}, where the adopted values are marked by
\(\dagger\).

\begin{table}[H]
\centering
\caption{Sensitivity analysis on Dataset~1.}
\label{tab:sensitivity_analysis}
\footnotesize
\setlength{\tabcolsep}{3.9pt}
\begin{tabular}{llccccc}
\toprule
\textbf{Parameter} & \textbf{Value} &
\textbf{F1}$\uparrow$ &
\textbf{FAR}$\downarrow$ &
\textbf{FNR}$\downarrow$ &
\textbf{Acc.}$\uparrow$ &
\textbf{Prec.}$\uparrow$ \\
\midrule

\multirow{4}{*}{Window length $L$}
& 5  & 0.9366 & 0.0115 & 0.0600 & 0.9814 & 0.9333 \\
& 10$^\dagger$ & \textbf{0.9535} & 0.0086 &
  \textbf{0.0432} & \textbf{0.9864} & 0.9502 \\
& 20 & \textit{0.9531} & \textbf{0.0084} &
  \textit{0.0450} & \textit{0.9863} & \textbf{0.9511} \\
& 30 & 0.9482 & 0.0092 & 0.0500 & 0.9848 & 0.9464 \\
\midrule

\multirow{3}{*}{Hidden dimension $d_h$}
& 32 & 0.9452 & 0.0102 & 0.0505 & 0.9839 & 0.9409 \\
& 64$^\dagger$ & \textbf{0.9535} & \textit{0.0086} &
  \textbf{0.0432} & \textbf{0.9864} & \textit{0.9502} \\
& 128 & \textit{0.9526} & \textbf{0.0083} &
  \textit{0.0462} & \textit{0.9861} & \textbf{0.9510} \\
\midrule

\multirow{4}{*}{$\tau_{\min}$}
& 0.05 & \textit{0.9514} & 0.0090 &
  \textit{0.0450} & \textit{0.9857} & \textit{0.9478} \\
& 0.10$^\dagger$ & \textbf{0.9535} & \textbf{0.0086} &
  0.0432 & \textbf{0.9864} & \textbf{0.9502} \\
& 0.50 & 0.9512 & 0.0100 &
  \textbf{0.0400} & 0.9856 & 0.9426 \\
& 1.00 & 0.9398 & 0.0092 &
  0.0660 & 0.9825 & 0.9456 \\
\bottomrule
\end{tabular}
\end{table}

The results reveal clear performance--complexity trade-offs. A short window (\(L=5\)) provides insufficient temporal context, whereas increasing the window beyond \(L=10\) yields only marginal changes and eventually degrades the overall detection balance. Similarly, \(d_h=32\) limits representation capacity, while \(d_h=128\) offers only minor improvements in FAR and
precision at the cost of greater complexity and slightly weaker F1-score, FNR, and accuracy. For the liquid dynamics, \(\tau_{\min}=0.10\) provides the most balanced operating point; smaller or larger values either offer no consistent improvement or reduce the detector's ability to respond effectively to changing evidence. Accordingly, we use \((L,d_h,\tau_{\min})=(10,64,0.10)\), which provides a good balance between temporal context, model capacity, responsiveness, and detection performance.

\subsection{Comparison with Reported GPS-IDS Results}

GPS-IDS~\cite{abrar2024gps} is selected as the closest published reference because it addresses the same GPS spoofing detection problem using the AV-GPS dataset family and vehicle-behavior information. We consider its Case~I results, in which the classifiers are trained on 80\% of Dataset~1 and evaluated on the remaining 20\% of Dataset~1 and on Dataset~3. This aligns the training-source direction with our evaluation, where the detector is developed using Dataset~1 before being transferred to Dataset~3.

The two methods nevertheless differ in their detection formulation. The published GPS-IDS evaluation applies conventional supervised classifiers to 14 time-indexed vehicle-behavior variables using stratified sample-level partitions. In contrast, the proposed method extracts physics-based evidence and tracks how it changes over time using only current and past observations, without mixing different trajectory segments. The comparison therefore provides performance context against the closest published framework rather than an identical sample-for-sample evaluation. Dataset~1 and Dataset~3 are reported because they constitute the primary in-domain benchmark and the continuous normal-to-attack transition evaluation, respectively. Dataset~2 is examined separately as a cross-location transfer experiment.

To avoid selecting different GPS-IDS classifiers for individual metrics, the classifier attaining the highest published F1-score on each dataset is used as the reference, and its complete metric vector is reported in Table~\ref{tab:gpsids_comparison}. F1-score is used for reference selection because it balances precision and recall.

\begin{table}[H]
\centering
\caption{Contextual comparison with the best-F1 published GPS-IDS
Case~I classifier on each primary dataset.}
\label{tab:gpsids_comparison}
\scriptsize
\setlength{\tabcolsep}{3.6pt}
\begin{tabular}{llcccc}
\toprule
\textbf{Data} & \textbf{Method} &
\textbf{F1}$\uparrow$ &
\textbf{Acc.}$\uparrow$ &
\textbf{Prec.}$\uparrow$ &
\textbf{Recall}$\uparrow$ \\
\midrule

\multirow{2}{*}{Dataset 1}
& Proposed
& \textbf{0.954}
& \textbf{0.986}
& 0.950
& \textbf{0.957} \\

& GPS-IDS (MLP)
& 0.948
& 0.974
& \textbf{0.979}
& 0.919 \\
\midrule

\multirow{2}{*}{Dataset 3}
& Proposed
& \textbf{0.978}
& \textbf{0.971}
& 0.983
& \textbf{0.973} \\

& GPS-IDS (RF)
& 0.972
& 0.965
& \textbf{0.997}
& 0.948 \\
\bottomrule
\end{tabular}
\end{table}

The proposed method achieves higher F1-score, accuracy, and recall on both datasets, whereas the selected GPS-IDS classifiers retain higher precision. This pattern indicates that the GPS-IDS references adopt more conservative positive decision behavior, while the proposed method provides stronger attack coverage and a better overall precision--recall balance. On
Dataset~1, the proposed method achieves a recall of 0.957, compared with 0.919 for GPS-IDS (MLP). On Dataset~3, it achieves a recall of 0.973, compared with 0.948 for GPS-IDS (RF), while also attaining the highest F1-score and accuracy.

The Dataset~3 result is particularly relevant because the proposed detector is transferred from Dataset~1 without retraining or threshold adjustment on the transition sequence. Its higher F1-score and recall are consistent with the intended benefit of causally integrating weak physical
inconsistencies over time. Because the published GPS-IDS results employ a different partitioning procedure, the table should be interpreted as a contextual literature comparison rather than an identical-protocol statistical ranking.

\section{Conclusion}

This paper proposed a causal high-order liquid evidence framework for detecting continuous and subtle GNSS spoofing attacks in autonomous driving. The framework constructs physics-guided evidence from the inconsistency between GNSS-implied and onboard-motion-derived displacements. It represents the residual level and its first- and second-order discrete variations as separate evidence streams and models their within-window evolution using separate adaptive liquid encoders. Hierarchical coupling progressively integrates the resulting order-specific temporal states, and the terminal fused representation
produces causal spoofing predictions using only current and past observations. Experiments on three AV-GPS subsets showed that the proposed method provides a strong balance between attack detection and false-alarm control on the in-domain and transition benchmarks, while detecting both labeled attack transitions within four sampling steps. The ablation results also confirmed the complementary roles of high-order residual evidence and adaptive liquid encoding. 

\begingroup
\small 
\bibliographystyle{IEEEtran}
\bibliography{ref}

@article{abrar2024gps,
  title={GPS-IDS: An anomaly-based GPS spoofing attack detection framework for autonomous vehicles},
  author={Abrar, Murad Mehrab and Youssef, Amal and Islam, Raian and Satam, Shalaka and Latibari, Banafsheh Saber and Hariri, Salim and Shao, Sicong and Salehi, Soheil and Satam, Pratik},
  journal={arXiv preprint arXiv:2405.08359},
  year={2024}
}

@article{dasgupta2022sensor,
  title={A sensor fusion-based GNSS spoofing attack detection framework for autonomous vehicles},
  author={Dasgupta, Sagar and Rahman, Mizanur and Islam, Mhafuzul and Chowdhury, Mashrur},
  journal={IEEE Transactions on Intelligent Transportation Systems},
  volume={23},
  number={12},
  pages={23559--23572},
  year={2022},
  publisher={IEEE}
}

@inproceedings{clements2022carrier,
  title={Carrier-phase and IMU based GNSS spoofing detection for ground vehicles},
  author={Clements, Zachary and Yoder, James E and Humphreys, Todd E},
  booktitle={Proceedings of the ION International Technical Meeting, Long Beach, CA},
  pages={83--95},
  year={2022}
}

@article{oligeri2022gps,
  title={GPS spoofing detection via crowd-sourced information for connected vehicles},
  author={Oligeri, Gabriele and Sciancalepore, Savio and Ibrahim, Omar Adel and Di Pietro, Roberto},
  journal={Computer Networks},
  volume={216},
  pages={109230},
  year={2022},
  publisher={Elsevier}
}

@article{national2021early,
  title={Early estimates of motor vehicle traffic fatalities and fatality rate by sub-categories in 2021},
  author={National Highway Traffic Safety Administration and others},
  journal={Crash• Stats Brief Statistical Summary. Report No. DOT HS},
  volume={813},
  pages={118},
  year={2021}
}

@article{wu2020spoofing,
  title={Spoofing and anti-spoofing technologies of global navigation satellite system: A survey},
  author={Wu, Zhijun and Zhang, Yun and Yang, Yiming and Liang, Cheng and Liu, Rusen},
  journal={IEEE Access},
  volume={8},
  pages={165444--165496},
  year={2020},
  publisher={IEEE}
}

@article{schmidt2020gps,
  title={A GPS spoofing detection and classification correlator-based technique using the LASSO},
  author={Schmidt, Erick and Gatsis, Nikolaos and Akopian, David},
  journal={IEEE Transactions on Aerospace and Electronic Systems},
  volume={56},
  number={6},
  pages={4224--4237},
  year={2020},
  publisher={IEEE}
}

@article{wang2023infrastructure,
  title={Infrastructure-enabled GPS spoofing detection and correction},
  author={Wang, Feilong and Hong, Yuan and Ban, Xuegang},
  journal={IEEE transactions on intelligent transportation systems},
  volume={24},
  number={12},
  pages={13878--13892},
  year={2023},
  publisher={IEEE}
}

@article{shafique2021detecting,
  title={Detecting signal spoofing attack in UAVs using machine learning models},
  author={Shafique, Arslan and Mehmood, Abid and Elhadef, Mourad},
  journal={IEEE access},
  volume={9},
  pages={93803--93815},
  year={2021},
  publisher={IEEE}
}

@article{namagembe2026machine,
  title={Machine Learning-Based GPS Spoofing Detection and Mitigation for UAVs},
  author={Namagembe, Charlotte Olivia and Ibrahim, Mohamad and Rahman, Md Arafatur and Pillai, Prashant},
  journal={Computers, Materials, \& Continua},
  volume={86},
  number={2},
  pages={1},
  year={2026},
  publisher={Tech Science Press}
}

@inproceedings{aissou2021tree,
  title={Tree-based supervised machine learning models for detecting GPS spoofing attacks on UAS},
  author={Aissou, Ghilas and Slimane, Hadjar Ould and Benouadah, Selma and Kaabouch, Naima},
  booktitle={2021 IEEE 12th Annual Ubiquitous Computing, Electronics \& Mobile Communication Conference (UEMCON)},
  pages={0649--0653},
  year={2021},
  organization={IEEE}
}

@article{yang2023anomaly,
  title={Anomaly detection against GPS spoofing attacks on connected and autonomous vehicles using learning from demonstration},
  author={Yang, Zhen and Ying, Jun and Shen, Junjie and Feng, Yiheng and Chen, Qi Alfred and Mao, Z Morley and Liu, Henry X},
  journal={IEEE Transactions on Intelligent Transportation Systems},
  volume={24},
  number={9},
  pages={9462--9475},
  year={2023},
  publisher={IEEE}
}

@article{radovs2024recent,
  title={Recent advances on jamming and spoofing detection in GNSS},
  author={Rado{\v{s}}, Katarina and Brki{\'c}, Marta and Begu{\v{s}}i{\'c}, Dinko},
  journal={Sensors},
  volume={24},
  number={13},
  pages={4210},
  year={2024},
  publisher={MDPI}
}

@article{bauer2025effect,
  title={Effect of GNSS Spoofing on GNSS-IMU Data Fusion-based Vehicle Pose Estimation},
  author={Bauer, Peter},
  journal={IFAC-PapersOnLine},
  volume={59},
  number={30},
  pages={150--155},
  year={2025},
  publisher={Elsevier}
}

@inproceedings{shen2020drift,
  title={Drift with devil: Security of $\{$Multi-Sensor$\}$ fusion based localization in $\{$High-Level$\}$ autonomous driving under $\{$GPS$\}$ spoofing},
  author={Shen, Junjie and Won, Jun Yeon and Chen, Zeyuan and Chen, Qi Alfred},
  booktitle={29th USENIX security symposium (USENIX Security 20)},
  pages={931--948},
  year={2020}
}

@article{dasgupta2022reinforcement,
  title={A reinforcement learning approach for global navigation satellite system spoofing attack detection in autonomous vehicles},
  author={Dasgupta, Sagar and Ghosh, Tonmoy and Rahman, Mizanur},
  journal={Transportation research record},
  volume={2676},
  number={12},
  pages={318--330},
  year={2022},
  publisher={SAGE Publications Sage CA: Los Angeles, CA}
}

@article{shabbir2023securing,
  title={Securing autonomous vehicles against gps spoofing attacks: A deep learning approach},
  author={Shabbir, Maliha and Kamal, Mohsin and Ullah, Zahid and Khan, Maqsood Muhammad},
  journal={IEEE Access},
  volume={11},
  pages={105513--105526},
  year={2023},
  publisher={IEEE}
}

@inproceedings{narain2019security,
  title={Security of GPS/INS based on-road location tracking systems},
  author={Narain, Sashank and Ranganathan, Aanjhan and Noubir, Guevara},
  booktitle={2019 IEEE Symposium on Security and Privacy (SP)},
  pages={587--601},
  year={2019},
  organization={IEEE}
}

@inproceedings{calvo2020lstm,
  title={LSTM-based GNSS Spoofing Detection Using Low-cost Spectrum Sensors},
  author={Calvo-Palomino, Roberto and Bhattacharya, Arani and Bovet, Gerome and Giustiniano, Domenico and others},
  booktitle={The 21st IEEE International Symposium on a World of Wireless, Mobile and Multimedia Networks (WoWMoM 2020)},
  year={2020}
}

@article{jiang2022deeppose,
  title={DeepPOSE: Detecting GPS spoofing attack via deep recurrent neural network},
  author={Jiang, Peng and Wu, Hongyi and Xin, Chunsheng},
  journal={Digital Communications and Networks},
  volume={8},
  number={5},
  pages={791--803},
  year={2022},
  publisher={Elsevier}
}

@inproceedings{dasgupta2023ai,
  title={Ai-based gnss spoofing attack detection for autonomous vehicles using satellite characteristics data},
  author={Dasgupta, Sagar and Rahman, Mizanur and Bandi, Thejesh N},
  booktitle={Proceedings of the 2023 International Technical Meeting of The Institute of Navigation},
  pages={514--525},
  year={2023}
}

@inproceedings{wen2023lstm,
  title={Lstm-based gnss spoofing detection for drone formation flights},
  author={Wen, Zheng and Qi, Xin and Sato, Toshio and Tamesue, Kazuhiko and Katsuyama, Yutaka and Sako, Kazue and Katto, Jiro and Sato, Takuro},
  booktitle={IECON 2023-49th Annual Conference of the IEEE Industrial Electronics Society},
  pages={1--6},
  year={2023},
  organization={IEEE}
}

@inproceedings{su2019robust,
  title={Robust anomaly detection for multivariate time series through stochastic recurrent neural network},
  author={Su, Ya and Zhao, Youjian and Niu, Chenhao and Liu, Rong and Sun, Wei and Pei, Dan},
  booktitle={Proceedings of the 25th ACM SIGKDD international conference on knowledge discovery \& data mining},
  pages={2828--2837},
  year={2019}
}

@inproceedings{audibert2020usad,
  title={Usad: Unsupervised anomaly detection on multivariate time series},
  author={Audibert, Julien and Michiardi, Pietro and Guyard, Fr{\'e}d{\'e}ric and Marti, S{\'e}bastien and Zuluaga, Maria A},
  booktitle={Proceedings of the 26th ACM SIGKDD international conference on knowledge discovery \& data mining},
  pages={3395--3404},
  year={2020}
}

@article{xu2021anomaly,
  title={Anomaly transformer: Time series anomaly detection with association discrepancy},
  author={Xu, Jiehui and Wu, Haixu and Wang, Jianmin and Long, Mingsheng},
  journal={arXiv preprint arXiv:2110.02642},
  year={2021}
}

@article{chen2021learning,
  title={Learning graph structures with transformer for multivariate time-series anomaly detection in IoT},
  author={Chen, Zekai and Chen, Dingshuo and Zhang, Xiao and Yuan, Zixuan and Cheng, Xiuzhen},
  journal={IEEE Internet of Things Journal},
  volume={9},
  number={12},
  pages={9179--9189},
  year={2021},
  publisher={IEEE}
}

@article{tuli2022tranad,
  title={Tranad: Deep transformer networks for anomaly detection in multivariate time series data},
  author={Tuli, Shreshth and Casale, Giuliano and Jennings, Nicholas R},
  journal={arXiv preprint arXiv:2201.07284},
  year={2022}
}

@inproceedings{liu2019secure,
  title={Secure pose estimation for autonomous vehicles under cyber attacks},
  author={Liu, Qipeng and Mo, Yilin and Mo, Xiaoyu and Lv, Chen and Mihankhah, Ehsan and Wang, Danwei},
  booktitle={2019 IEEE Intelligent Vehicles Symposium (IV)},
  pages={1583--1588},
  year={2019},
  organization={IEEE}
}

@inproceedings{he2019temporal,
  title={Temporal convolutional networks for anomaly detection in time series},
  author={He, Yangdong and Zhao, Jiabao},
  booktitle={Journal of Physics: Conference Series},
  volume={1213},
  number={4},
  pages={042050},
  year={2019},
  organization={IOP Publishing}
}

@article{wu2022timesnet,
  title={Timesnet: Temporal 2d-variation modeling for general time series analysis},
  author={Wu, Haixu and Hu, Tengge and Liu, Yong and Zhou, Hang and Wang, Jianmin and Long, Mingsheng},
  journal={arXiv preprint arXiv:2210.02186},
  year={2022}
}

@article{chen2018neural,
  title={Neural ordinary differential equations},
  author={Chen, Ricky TQ and Rubanova, Yulia and Bettencourt, Jesse and Duvenaud, David K},
  journal={Advances in neural information processing systems},
  volume={31},
  year={2018}
}

@article{rubanova2019latent,
  title={Latent ordinary differential equations for irregularly-sampled time series},
  author={Rubanova, Yulia and Chen, Ricky TQ and Duvenaud, David K},
  journal={Advances in neural information processing systems},
  volume={32},
  year={2019}
}

@inproceedings{hasani2021liquid,
  title={Liquid time-constant networks},
  author={Hasani, Ramin and Lechner, Mathias and Amini, Alexander and Rus, Daniela and Grosu, Radu},
  booktitle={Proceedings of the AAAI conference on artificial intelligence},
  volume={35},
  number={9},
  pages={7657--7666},
  year={2021}
}

@article{hasani2022closed,
  title={Closed-form continuous-time neural networks},
  author={Hasani, Ramin and Lechner, Mathias and Amini, Alexander and Liebenwein, Lucas and Ray, Aaron and Tschaikowski, Max and Teschl, Gerald and Rus, Daniela},
  journal={Nature Machine Intelligence},
  volume={4},
  number={11},
  pages={992--1003},
  year={2022},
  publisher={Nature Publishing Group UK London}
}

@IEEEtranBSTCTL{BSTcontrol,
  CTLdash_repeated_names = "no"
}

@article{dasgupta2020prediction,
  title={Prediction-based GNSS spoofing attack detection for autonomous vehicles},
  author={Dasgupta, Sagar and Rahman, Mizanur and Islam, Mhafuzul and Chowdhury, Mashrur},
  journal={arXiv preprint arXiv:2010.11722},
  year={2020}
}
\endgroup

\end{document}